\documentclass[conference]{IEEEtran}
\usepackage[style=ieee]{biblatex}
\usepackage{amsmath,amssymb}
\usepackage{authblk}
\usepackage{upgreek}
\usepackage{pgfplots}
\usepackage{filecontents}
\usepackage{graphicx,subcaption}
\usepackage{autobreak}

\usepackage{siunitx}
\usepackage{tikz}
\usepackage[section]{placeins}

\pgfplotsset{compat=newest} 
\usepgfplotslibrary{units} 

\DeclareMathOperator{\EX}{\mathbb{E}}
\usepackage{times}
\usepackage{helvet}
\bibliography{Research_Paper}
\usepackage{courier}
\usepackage{amsmath}
\author[]{Deepak-George Thomas}
\author[]{Daniil Olshanskyi}
\author[]{Karter Krueger}
\author[]{Tichakorn Wongpiromsarn
}
\author[]{Ali Jannesari}

\affil[]{Department of Computer Science, Iowa State University, USA\\\{dgthomas, daniilo, karterk, nok, jannesari\}@iastate.edu}
\begin{document}

\title{Interpretable UAV Collision Avoidance using Deep Reinforcement Learning}
\maketitle
\begin{abstract}
\begin{quote}
The significant components of any successful autonomous flight system are task completion and collision avoidance. Most deep learning algorithms successfully execute these aspects under the environment and conditions they are trained. However, they fail when subjected to novel environments. This paper presents an autonomous multi-rotor flight algorithm, using Deep Reinforcement Learning augmented with Self-Attention Models, that can effectively reason when subjected to varying inputs.  In addition to their reasoning ability, they are also interpretable, enabling it to be used under real-world conditions.  We have tested our algorithm under different weather conditions and environments and found it robust compared to conventional Deep Reinforcement Learning algorithms.
\end{quote}
\end{abstract}

\section{Introduction}

The use of AI-powered drones for various tracking problems has become widely prevalent. They have been recently used for fighting forest fires, searching for a target during disasters and observing wildlife. In addition, recent work has focused on developing autonomous online algorithms that control the trajectory of the Unmanned Aerial Vehicle (UAV) under different mission scenarios. The ability of UAVs to adapt to diverse and novel constraints comes from the path planning capability of the underlying algorithm controlling it. \cite{hayat2017multi, jiang2018online, haksar2018distributed, wu2019uav, lyons2019monitoring, asli2019energy}. Deep Reinforcement Learning (DRL) is usually preferred for this task over the conventional approaches such as A*, artifical potential fields, Floyd warshall, dynamic programming and similar methods, since it does not present the same disadvantages as other methods which make it infeasible to implement in real world situations, \cite{radmanesh2018overview, bounini2017modified, sathyaraj2008multiple, lamini2018genetic, villanueva2019deep, vaddi2019efficient, mammadli_ea:taco:2019}. \\

Reinforcement Learning (RL) is associated with an agent or a set of agents, placed within an environment, that take a series of actions to maximize their expected reward \cite{franccois2018introduction}. A Model Free RL algorithm, Q-Learning, was developed to obtain the action value for specific states \cite{watkins1993technical}. Deep Q-Learning (DQN) (a class of DRL algorithms) combined Q-Learning with a Deep Neural Network (DNN) \cite{mnih2015human}. A DQN consists of multiple hidden layers, thereby allowing it to learn and function well in situations involving high dimensional spaces. Therefore, it can mitigate to some extent, the curse of dimensionality problem affecting Q-Learning \cite{arulkumaran2017deep}.\\
Furthermore, the training of robots under real-world conditions is an expensive and time-consuming process. Therefore, it is beneficial to train the algorithm under realistic simulators and then transfer the parameters learned by the algorithm from the system into a real-world robot. Since simulated training cannot capture all the aspects and nuances of the real world, it is another reason why the training needs to be robust to novel environments and changes in weather conditions and lightning. A further motivation for the algorithm to be generalizable is that using all aspects of the real world is not possible in simulation. Therefore, we can only train it with a subset of such features. \\
Our objective is to improve the learning and generalization capability of our multirotor using self-attention models. For collision avoidance tasks, a Convolutional Neural Network (CNN) is generally used as a function approximator. The CNN uses the robot camera as input and these inputs are then considered to be the ‘State’ in the DQN algorithm. The output of the CNN is a set of Q-values equaling the number of actions. Based on the Q value, the algorithm chooses an action that is then used to maximize the expected reward \cite{roghair2021vision}. The disadvantage of using CNNs is that they are generally not invariant to deformations or rotations. Therefore, the learning algorithm may not be able to classify, similar objects within the environment under the same category if they have been deformed or modified in any way. To enable the DNN to generalize to modifications in the environment, it needs to have relational properties, which allows it to classify objects while relating them with other objects \cite{zai2020deep, zambaldi2018relational}. Hence, we use a Self-Attention Model with Duelling Deep Q Network (D3QN-SAM) for collision avoidance, while comparing its performance with a Duelling Deep Q-Network that uses a CNN (D3QN-CNN). 
\section{Related Work}
\subsection{Attention Augmented Reinforcement Learning}
\cite{zambaldi2018relational} borrowed ideas from the Relational RL umbrella (RRL) \cite{dvzeroski1998relational, dvzeroski2001relational} that was introduced more than 20 years prior. RRL uses relational representations to represent states, policies and actions, and combines it with RL. In addition, they implemented a self attention model to compute non-local interactions. Their basis for choosing this model was that it wasn't computationally intensive and shared features with graph neural networks. Furthermore, this model enabled obtaining relations between different entities. Therefore they incorporated relational reasoning coupled with structured perception to develop RL algorithms capable of interpretable reasoning. It could be discerned that agents trained using such architectures tend to be more generalized and providing better performance compared with baseline models. Also, they found that the algorithm's internal computations could be visualized and the computations were in line with those architectures that obtained task-specific relations \cite{zai2020deep}.\\
\cite{goyal2019recurrent} developed a novel architecture consisting of various recurrent cells that had almost independent transition dynamics, known as Recurrent Independent Mechanism (RIM). Similar to \cite{zambaldi2018relational}, they used multiple attention heads that enabled communication between RIMs. In addition, these attention heads were used to select which RIMs were required for the input at a particular time step. They claimed that usually numerous processes consist of sub-processes working independently. However, ML models assume that all of these processes interact degrading the generalization of these algorithms. Therefore, to address this issue, they used RIMs that contended for attention during all time steps and were updated at only pertinent steps. The authors demonstrated that this technique improved the generalization capability across a wide range of tasks such as those having temporal patterns, objects and a combination of patterns and objects.\\ 
\cite{tang2020neuroevolution} extended the idea of self-attention for visual RL tasks. They use neuro-evolution to train their DRL agent that used self-attention to focus on critical aspects within images in order to finish their tasks. It was observed that the agent's focusing ability enables it to generalize to new environments, since it concentrated on few (necessary) aspects of the visual stimuli and thereby was able to beat conventional algorithms with relatively lower parameters. However, the authors noted that their algorithm still couldn't compete with human abilities of generalization since despite the change in environment, the authors kept the properties of objects obtaining high attention the same. They claimed that modifying those properties could lead to performance degradation.\\
\subsection{Collision Avoidance using Reinforcement Learning}
\cite{hodge2021deep} used DRL coupled with Long Short Term Memory Networks to assist drone navigation around city environments. Their algorithm was used to assist pilots under safety critical scenarios. In addition, they implemented curriculum learning to augment their algorithm, wherein training starts on a single task whose complexity is increased as learning progresses \cite{bengio2009curriculum}. However, the use of attention based mechanisms including transformers have been used to replace LSTMs and GRUs due to their inherent limitations  \cite{vaswani2017attention}.\\
Moreover, \cite{roghair2021vision} developed an autonomous multi-rotor flight DRL algorithm that enabled collision avoidance under novel scenarios while encouraging exploration. The algorithm used depth images as input while flying across an enclosed space with the intention of maximizing the number of steps taken before a crash. Since the rewards obtained for this task was sparse,  they developed a guidance based approach to motivate their agent to explore \cite{gou2019dqn}. In addition to the existing standard D3QN networks (online and target), an additional CNN network was used to predict the future state and this prediction was then compared to the states in the replay memory. Thereby, they rewarded the agent if it moved to states that had least similarity with the ones already present in the replay memory. A Bayesian Gaussian Mixture Model was used to obtain a similarity value between the states. This technique enabled the agent to explore the environment and address the sparsity in rewards. The shortcoming of this algorithm is that the Gaussian Mixture Model used for assessing similarity is computationally costly and training would take a long time for large environments.\\
Also, \cite{singla2019memory} used Deep Recurrent Q-Networks augmented using Temporal attention for the purpose of collision avoidance in multi-rotors. In addition, they use Conditional Generative Adversarial Neural Networks to estimate depth images from inputs obtained using a monocular depth camera. This algorithm was tested  in a simulated room full of people and their algorithm was able to beat conventional baselines such as DQN and D3QN. A potential drawback of this algorithm is that \cite{hausknecht2015deep} experimentally showed that the addition of recurrent layers did not help improve the performance of the algorithm relative to stacking observations while using a CNN. Therefore, stacking multiple observations together before passing it through a CNN could make the use of Recurrent Q-Networks redundant while reducing the computational overhead obtained from using RNNs.\\
\section{Background}
\subsection{Duelling Double Deep Q Networks}
Our work deals with a single RL agent that has to make sequential decisions in discrete time steps while interacting with its environment, $\epsilon$. The state space, $S$ contains M images, $s_t = (x_{t-M+1},..., x_{t})$, where $t$ refers to the time step. Also, we take our action space, $A$ to be discrete, $a_t = {1,...,\mid{A}\mid}$ and depending upon the action the agent obtains a reward $r_t$. As previously described, the agent in a RL algorithm attempts to maximize its expected return. In addition, the return is subject to discounting, $\gamma$ which is used to determine the value difference between future and current rewards. \\
$R_t = \sum_{\tau = t}^{\infty} \gamma^{\tau-1}r_{\tau}$, where $\gamma \in [0,1]$ \\
The value of being in state $s$ or the state-action $(s,a)$, provided the agent follows the policy $\pi$ is given by - \\
$Q^{\pi}(s,a) = \mathbb{E}[R_t|s_t = s, a_t = a, \pi]$\\
$V^{\pi}(s) = \mathbb{E}_{a\sim\pi(s)}[Q^{\pi}(s,a)]$\\
Dynamic programming can be used to compute the following Q function - \\
$Q^{\pi}(s,a) = \mathbb{E}_{s'}[r + \gamma\mathbb{E}_{a'\sim\pi(s')}[Q^{\pi}(s',a')]|s,a,\pi]$\\
The optimal $Q^{*}(s,a)$ is given by, $Q^{*}(s,a) = max_{\pi}Q^{\pi}(s,a)$. Also, we get $V^{*}s = max_{a}Q^{*}(s,a)$ from a deterministic policy,\\ $a = \arg \max_{a' \in A}Q^{*}(s,a')$. Therefore the optimal Q function can be obtained as follows -\\
$Q^{*}(s,a) = \mathbb{E}_{s'}[r + \gamma max_{a'}[Q^{*}(s',a')]|s,a]$\\
Moreover, based on the above equations, a new function can be defined that is used to associate the Q and value functions;\\
$A^{\pi}(s,a) = Q^{\pi}(s,a) - V^{\pi}(s)$, where $A$ is the advantage function\\ From the preceding equations, it is easy to discern that the ,\\ $\mathbb{E}_{a~\pi(s)}[A^{\pi}(s,a)] = 0$. The relative significance of each action can be found using the advantage function, since it is the difference of the value of taking a specific action within a state and the value of being in that state.  \\
To approximate the value function, a deep Q-network can be employed: $Q(s, a; \theta)$ having parameter $\theta$. Training the DNN involves minimizing the following loss function;\\
$L_{i}(\theta_{i}) = \EX_{s,a,r,s'}[(y_{i}^{DQN}-Q(s,a;\theta_i))^2]$\\
where $y_{i}^{DQN} = r + \gamma*max_{a'}Q(s',a';\theta^{-}),$ and $i$ is the iteration at that time step. Here $\theta^{-}$ constitutes the online and target network parameters. Having two networks, instead of one, has been shown to improve the stability of the DQN. As per this technique, the online network is updated every iteration, however the target network is allowed to update its parameters  only at specific iterations \cite{mnih2015human}. \\
The downside of both DQN and Q-learning is that the values used to choose and evaluate an action are the same, leading to value estimates that are overoptimistic. Double Deep Q-Networks  addressed this issue by replacing the target $y_{i}^{DDQN}$ with $y_{i}^{DQN}$ as given below; \cite{van2015deep}\\
$y_{i}^{DDQN} = r + \gamma Q(s',arg_{a'}max(s',a';\theta_{i});\theta^{i})$ \cite{hasselt2010double} \\
The motivation behind the Dueling Network Architecture was the observation that calculating the action values for various states did not help the agent. The authors demonstrated this while playing Enduro, the knowledge regarding left or right movements did not affect the game unless there was a chance of collision . The architecture consisted of a Q-network called the duelling network. The lower Convolutional layers of the dueling network were then split up into two fully connected layers estimating the advantage and value functions. Ultimately, the two layers were joined to output a Q function consisting of Q values for every action. \cite{mnih2015human} \\
Furthermore, the design of the Dueling Network Architecture consisted of a few intricate details. We know that $Q^{\pi}(s,a) = V^{\pi}(s)+A^{\pi}(s,a)$ and $V^{\pi}(s) = \mathbb{E}_{a \sim\pi(s)}[Q^{\pi}(s,a)]$, we can infer that \\$\mathbb{E}_{a \sim \pi(s)}[A^{\pi}(s,a)] = 0$\\ Also, in the case of a deterministic policy, $a^{*} = \arg \max_{a' \in A} Q(s,a')$, we can infer that $Q(s,a^{*}) = V(s)$ leading to $A(s,a^{*}) = 0$
The two fully connected layers in the dueling network yield a scalar $V(s;\theta,\beta)$ and the $\mid{A}\mid- $ dimensional vector $A(s,a;\theta, \alpha)$ respectively. Here $\alpha, \beta$ represent the parameters of the fully connected layers while $\theta$ denotes the convolution  parameters.\\
Moreover, the authors claimed that inferring the following equation from the above steps would be wrong;\\
$Q(s,a;\theta, \alpha, \beta) = V(s; \theta, \beta) + A(s,a;\theta, \alpha)$\\
This is because $V(s; \theta, \beta)$ and $A(s,a;\theta, \alpha)$ might not be good estimates of their functions. In addition, $Q(s,a;\theta, \alpha, \beta)$ is not the Q-function but only an estimate of it. Also, the authors found that the above equation is unidentifiable, since one cannot find unique values of $V$ and $A$ if they have $Q$. This problem can be solved by coercing the estimator of the advantage function at the specified action to provide no advantage. Therefore, the authors ran forward mapping using the network's final module leading to the advantage function estimator obtaining a zero advantage for the specific action. \\
$Q(s,a;\theta,\alpha, \beta) = V(s;\theta, \beta) + (A(s,a;\theta,\alpha) -\\max_{a' \in |A|}A(s,a';\theta, \alpha))$\\
Also, from $a^{*} = \arg \max_{a' \in A}Q(a,a';\theta, \alpha, \beta) = \\ \arg \max_{a' \in A} A(s,a';\theta,\alpha)$ leads to $Q(a, a^{*};\theta, \beta) = V(s;\theta,\beta)$. Therefore from the layers $A(s,a;\theta, \alpha)$ we get an estimate of the advantage function whereas from the layers $V(s;\theta,\beta)$ we get the value function estimation. \\
In addition, to increase the stability of optimization the authors replaced the max operator with average;\\
$Q(s,a;\theta,\alpha,\beta) = V(s,\theta,\beta) + (A(s,a;\theta,\alpha)-1/{|A|}\sum_{a'}A(s,a';\theta,\alpha))$\\
Finally, it was noted that learning algorithms using Q networks in general could now be trained on the dueling architecture. 
\cite{wang2016dueling}
\subsection{Self-Attention Models}
Self-Attention Models (SAM) can be used to obtain attention weights among all nodes within a set, $N:\mathbb{R}^{n \times f}$. Here $n$ and $f$ represent the number of nodes and the features associated with it respectively, The result of a SAM can be understood as an edge matrix with dimension $n \times n$, $E: \mathbb{R}^{n \times n}$. The edge matrix enables the observer to discern the relation between various nodes leading to a better understanding of how the DNN "reasons". Nodes that are highly related can communicate and therefore update their parameters. This entire process takes place within a single step and is referred to as the relational module. The output of a relational module, $N:\mathbb{R}^{n \times f}$, is then sent to the next relational module.  \\
SAM have been shown to work with great success on language models in the past. However, in recent years it has also been used and demonstrated great promise on vision tasks \cite{ramachandran2019stand, Zhao_2020_CVPR}. However, the visual data needs to be pre-processed before it can be used as input within a relational module. In essence, in order to represent images as nodes we need to convert pixel data into objects.To achieve this, a CNN is used to convert the input images into a collection of feature maps. The feature maps are responsible for learning fine and rough features that are present within the images. These features are then concatenated together along its channels thereby creating a cuboid. This cuboid can then be sliced along the horizontal and vertical dimensions to create objects that can be viewed as nodes of a graph which is used as input to the relational module.
\\
The relational module architecture described below was presented in the paper by \cite{zambaldi2018relational}. The node matrix, $N:\mathbb{R}^{n \times f}$ is projected into key, query and value matrices. The matrices responsible for projecting the initial node matrix are set as learnable parameters during training and should converge to those than can generate optimal attention weights from the key, value and query matrices. \\
To facilitate a better understanding, the SAM for two nodes, $a:\mathbb{R}^{10}$ and $b:\mathbb{R}^{10}$ has been described here. We get key, value and query matrices from these nodes through their projection matrices. \\
$a_Q = a^{T}Q$, $a_K = a^{T}K$, $a_V = a^{T}V$\\
To obtain the self similarity for the nodes, an inner product operation is performed between the key and query matrices of the respective nodes. Generally the similarity measure for a node with itself is expected to be high. However, depending upon its key and query that might not be the case. An un-normalized scalar attention weight is generated by the product of the key and query matrices of a particular node, $w_{a,a}=<a_{Q}, a_{K}>$. This process can now be extended to both of the nodes a and b. This leads to four attention weights, (two for the interaction between the nodes and two for interaction with themselves). A softmax function is then used to normalize the attention weights so that each of them are between $[0,1]$ and they sum to 1. This normalization is crucial to ensuring that the attention weight focus only on the necessary nodes. \\
The attention weights for the nodes a and b are stored in a $2 \times 2$ matrix. Subsequently, the value matrix gets updated by the product of the attention weights and the value matrix. Therefore, each node is modified depending upon its connection with the other. This process can now be generalized to $n$ nodes. \\
The process of generating an attention matrix by multiplying the key and query together and then using it to update the value matrix along with normalization can be combined and represented as a matrix operation - \\
$\hat{N} = softmax(QK^{T})V$\\
Here n, f, Q, K and V represent the number of nodes, node feature vector dimension, query matrix, key matrix and value matrix respectively with the following dimensions; $Q:\mathbb{R}^{n \times f}$, $K^{T}:\mathbb{R}^{f \times n}$ and $V:\mathbb{R}^{n \times f}$. The attention weight matrix $(QK^{T})$ will have dimension $n \times n$ with the value at $(x,y)$ demonstrating the strength of the interaction of node $x$ with the node at $y$. The product of the normalized attention weight matrix, $A$ with the value matrix updates the feature vector of all nodes corresponding to its interaction with others, leading to a node matrix with dimensions; $\hat{N}:\mathbb{R}^{n \times f}$. The last step to create a relational block is to run a  linear layer over the resulting node matrix before implementing non-linearity for augmenting the learning process. Multiple relational blocks can be stacked together to increase the model complexity leading to better learning of higher order relations \cite{zai2020deep, vaswani2017attention}.\\
\subsection{Multi-Head Self Attention}
The use of softmax in SAM models restricts the the attention weights that can be placed on individual objects, especially for cases where there are a large number of nodes. Moreover, this leads to cases wherein the attention is being placed on a limited number of nodes and others are being ignored. In order to mitigate the amount of information lost, we can use multiple independent attention heads and subsequently concatenate them together. Therefore, these attention heads could concentrate on different areas of the image, thereby increasing the amount of learning without affecting the relational modules. Another advantage of attention heads is that nodes corresponding to respective heads can concentrate on a smaller set of nodes, instead of providing a minute amount of weights by trying to cover all the nodes, leading to increased interpretability  \cite{zai2020deep, vaswani2017attention}.

\section{Problem Description}
\subsection{Problem Statement}
Our goal is to train a multi-rotor to avoid obstacles while flying around an area using DRL algorithms augmented with SAMs and CNNs. In addition, we test the performance of our trained algorithms on different weather conditions as well as new environments. Finally, we visualize the attention weights in order to interpret the decisions made by our algorithm.
\subsection{Microsoft Airsim}
\begin{figure}[htp]
    \centering
    \includegraphics[width=8cm]{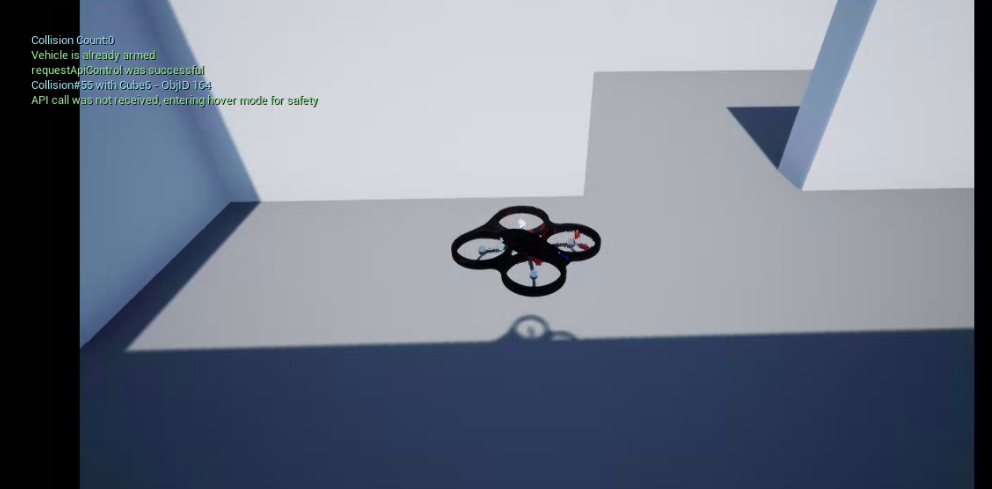}
    \caption{Training Environment}
    \label{fig:env}
\end{figure}

The simulation environment was generated using the Microsoft Airsim Simulator, which provided the physics behind the control of the UAV Quadcopter used in the project. It can be implemented like a plugin for Unreal and can work with all Unreal environments. This is beneficial for robotic applications such as collision avoidance since evaluating such algorithms in the real world is expensive. Furthermore, Airsim can accommodate UAV movements such as pitch, roll and yaw through its API calls. In addition, the communication between the python RL program and the simulation occurs through the Airsim API. Figure~\ref{fig:env}, below shows the 3-D environments developed for training our agent using the Unreal Engine \cite{zuluaga2018deep, shah2018airsim}. 

\subsection{MDP Formulation}
\subsubsection{State Space}
\begin{figure}[htp]
    \centering
    \includegraphics[width=8cm]{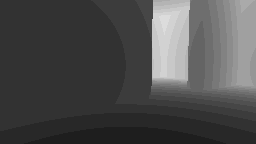}
    \caption{Image from Depth Camera}
    \label{fig:galaxy}
\end{figure}

The state space $S$ for our algorithm is in the form of depth images taken by the depth camera during training in Airsim (Figure~\ref{fig:galaxy}). Similar to \cite{mnih2015human}, we stack 5 frames together to model the sequential movement of the drone.
\subsubsection{Action Space}
One of the following four actions is executed by the agent during every time step.\\
$A= \{a_0, a_1, a_2, a_3\}$\\
The linear velocity $(v)$ is controlled by two actions, $a_0$ and $a_3$ with a speed of 1m/s and 2m/s respectively. The actions $a_1$ and $a_2$ consist of a combination of angular and linear velocity. $a_1$ makes the multi-rotor take a \ang{30} yaw-right and then move in the forward direction at 1m/s. Similarly,  $a_2$ forces the agent to take a \ang{30} yaw-left and then move forward at 1m/s.
\subsubsection{Reward Function}
The reward function was designed so that the UAV learns to avoid collisions while staying away from walls and objects. Therefore it consists of two major components. The first component gives it a negative reward of -10 in case it collides against the wall. However, minor brushes against the wall are ignored. The next component rewards the UAV for staying away from objects. At each time step, it extracts the centre of the image feed using $20 \%$ of the width and height of depth frame. Next, it finds the minimum value within that region $\beta$ and compares it against a minimum distance threshold, $\delta = 3$. If the minimum value of the center of the depth image is greater than the threshold, we consider the UAV to be in a safe region and reward it proportionally based on how far it is from the wall. Therefore under this condition, the agent gets a reward of $3 + \beta/5$. This gives the agent a reward of 3 for maintaining that threshold along with a proportionally increasing reward for finding the "most-open" area centered in front of it. For cases where $\beta$ is less than $\delta$, a negative reward proportional to how close it is to the walls is placed on the agent, equalling $\delta/(\beta+0.5)*-1$ . We add a $0.5$ to the denominator to account for cases where $\beta = 0$, when the multi-rotor is extremely close to the wall, and prevent the expression from becoming $-\inf$.\\
\begin{equation*}
    R = \begin{cases}
    \delta/(\beta+0.5)*-1 &\text{$\beta<\delta$} \\
    3 + \beta/5 &\text{$\beta>\delta$} \\
    -10 &\text{Collision Detected} 
    \end{cases}
    \end{equation*}
\subsubsection{Model Dynamics}
The reinforcement learning model largely follows work done by \cite{zambaldi2018relational}. It takes in depth images as input and calculates the $Q(s_{t}, a_{t})$ for all actions while it is at the state $s_{t}$. After taking an action based on $\arg \max_{a}Q(s_t, a_t)$ it reaches the next state $s_{t+1}$. The tuple containing the information $(s_t, a_t, r_t, s_{t+1})$ are stored in the replay memory $M$. In the event of a crash, the environment resets and the UAV returns to its initial position. The model uses a CNN to learn features obtained from the simulation depth camera. Next, it converts these features along the channel dimension to form individual nodes. This converts our two-dimensional image into a graph-based representation. Moreover, it projects these nodes into key, query and value matrices. This representation enables the attention algorithm to compute the relation between different nodes and generate an attention matrix. The last steps include running the attention matrix through linear layers and applying pooling operations (similar to that of CNN operations) to generate an output layer with nodes equaling the number of actions of the agent.

\section{Results and Discussions}
\subsection{Training Results}
\begin{figure}[!ht]
    \centering
    \includegraphics[width=8cm]{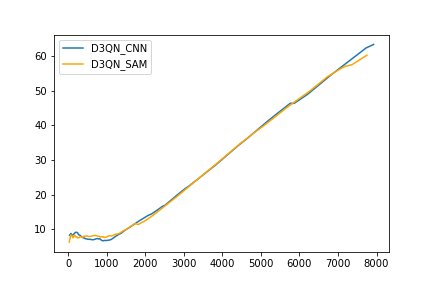}
    \caption{Average Episode Length}
    \label{ep_len}
\end{figure}

\begin{figure}[!ht]
    \centering
    \includegraphics[width=8cm]{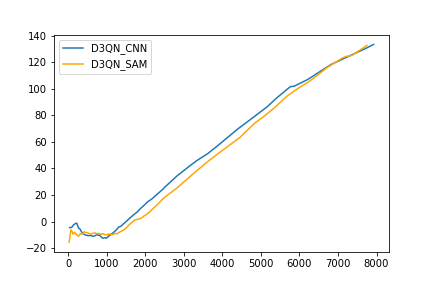}
    \caption{Average Episode Reward}
    \label{ep_rew}
\end{figure}

In order to compare the performance of the SAM model with a conventional baseline, we trained two D3QN agents. One agent had a SAM (with 3 heads) as its function approximator whereas the other had a CNN. Each agent was trained for a total of 8000 epochs. While tuning the number of epochs, we found that using a large number of epochs led to catastrophic forgetting \cite{kirkpatrick2017overcoming}, wherein the agent's performance would suddenly drop and it seemed to 'forget' whatever it had learned during previous iterations. Furthermore, the training metrics are calculated based on episodes, which can be defined as the series of steps between each episode reset. The average episode length (Figure ~\ref{ep_len}) and the average reward (Figure ~\ref{ep_rew}) obtained by the agent was used to discern how well the learning process was going. The D3QN SAM and the D3QN CNN have quite similar plots. Also, the mean episode length and episode reward follow the same trend, which implies that that reward function is helping the agent with its objective. Moreover, during the epsilon-decay process, wherein the $\epsilon$ is reduced during the course of training, the final epsilon was kept at 0.05. This makes the agent take rare, random actions even after the $epsilon$ has completely decayed.  \\
Furthermore, we found that it was important to view the training videos in simulation. Initially, we set our actions $a_1$ and $a_2$ to consist only of yawing actions minus the linear velocity. The agent increased its rewards and steps just by staying at one place and rotating indefinitely. \\
While comparing our work with \cite{roghair2021vision}, we noted that the training times using the Guidance Training algorithm were a lot larger and that it took days to train before it attained a reasonable policy. In contrast, the SAM model was able to consistently increase its rewards and reduce its crash rate in a few hours. Furthermore, \cite{singla2019memory} used temporal attention with DRL that allocates weights to past observations(images) based on its importance and then predicts a trajectory while faced with moving obstacles. However, our SAM algorithm instead detects the relationship between different objects in a single image, which allows the algorithm to be robust in the presence of noise and ,to a certain extent, invariant to changes in the obstacles geometry \cite{pei2017temporal}.

\subsection{Testing Results}
\begin{figure}[!ht]
    \centering
    \includegraphics[width=8cm]{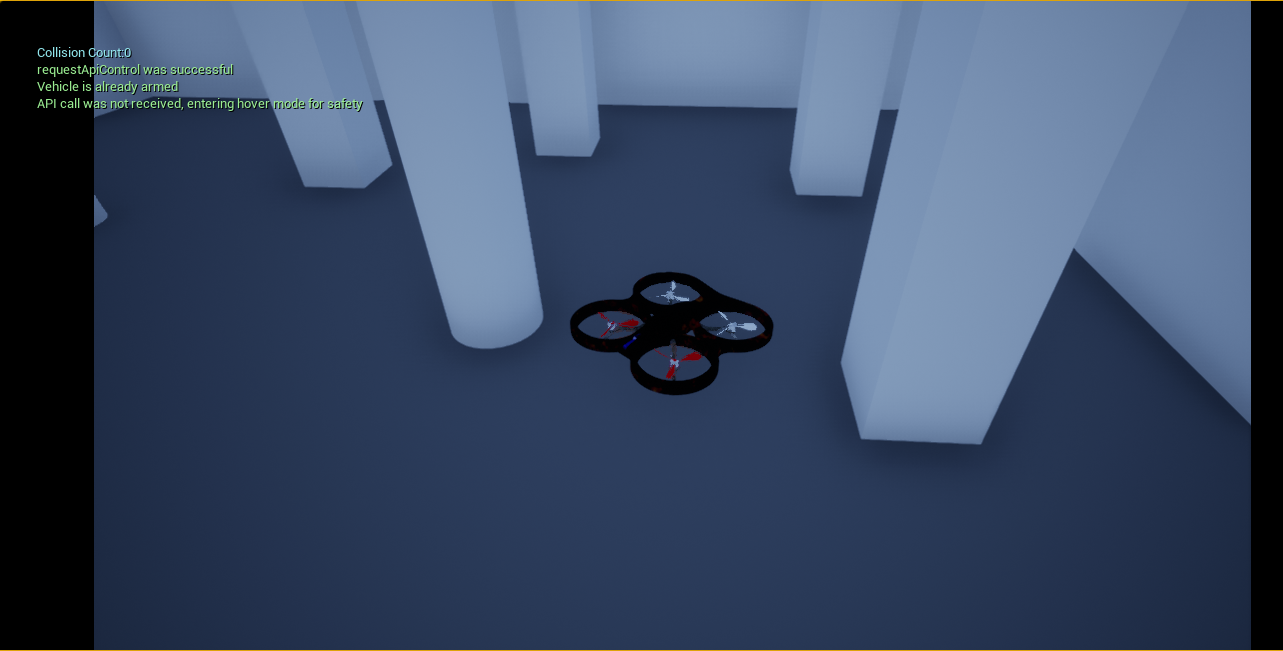}
    \caption{Complex Environment}
    \label{Complex_Environment_2}
\end{figure}

\begin{figure}[!ht]
    \centering
    \includegraphics[width=8cm]{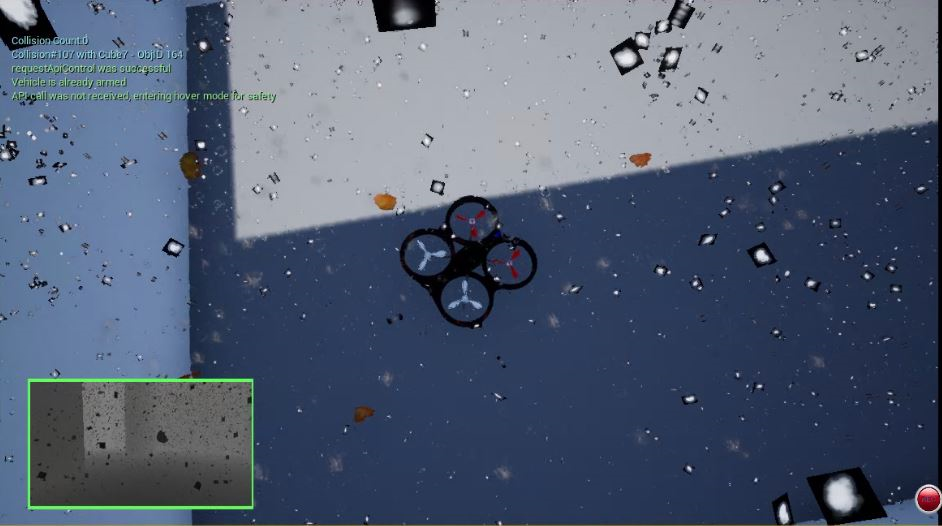}
    \caption{Turbulent Environment}
    \label{weather}
\end{figure}

To test the robustness of our algorithm, we added various obstacles as well tested the performance of the algorithm under different weather conditions including rain, snow and falling leaves for 1000 time steps. During this evaluation, the parameters were fixed and the D3QN algorithms were only allowed to exploit without doing any exploration. The algorithms were first tested on the environment in which they were trained (Basic Environment). Next, we developed a new environment consisting of cylindrical and rectangular beams as shown in Figure~\ref{Complex_Environment_2} and referred to it as the Complex Environment. Finally, we changed the weather conditions within the Basic environment using a combination of snow, rain and leaves and called it the Turbulent Environment (Figure~\ref{weather}). The results of the testing process are shown in Table ~\ref{table:testing_results}.\\
\begin{table}[!ht]
\caption{Number of Crashes during Testing} 
\centering 
\begin{tabular}{c c c} 
\hline\hline 
Case & D3QN-SAM & D3QN-CNN \\ [0.5ex] 
\hline 
Basic Environment & 0 & 8\\ 
Complex Environment & 1 & 5 \\
Turbulent Environment & 1 & 10 \\
[1ex] 
\hline 
\end{tabular}
\label{table:testing_results} 
\end{table}
It is evident from Table ~\ref{table:testing_results} that the D3QN-SAM performs much better than the the D3QN-CNN under all cases. The D3QN-SAM had 0 failures in the Basic Environment, while the D3QN-CNN failed 9 times which is surprising as both algorithms had similar results during training. The architecture of the Complex Environment was drastically different from the Basic Environment for we wanted to check whether our algorithms actually learned a collision avoidance policy rather than just follow a fixed path that minimized contact with obstacles in the training environment. In addition, this environment was filled with beams and the agent would have to constantly make detours to avoid crashing. Based on the low number of crashes we can confidently claim that the algorithms learned the desired policy with the D3QN-SAM having just 1 failure in this environment. The Turbulent Environment was designed so as to distort the camera feed and therefore the input states provided to the DRL algorithm. The change in the depth image sent to the algorithm can be viewed at the lower left of Figure~\ref{weather}. The D3QN-SAM was robust to change in this feed for, based on its relational modules, it could identify the relation between objects and place attention on the pertinent aspects of the state. 
\begin{table}[!ht]
\caption{Average Rewards during Testing} 
\centering 
\begin{tabular}{c c c} 
\hline\hline 
Case & D3QN-SAM & D3QN-CNN \\ [0.5ex] 
\hline 
Basic Environment & 2.406
 & 2.459
\\ 
Complex Environment & 2.16
 & 2.319
 \\
Turbulent Environment & -7.809
 & -7.83
 \\
[1ex] 
\hline 
\end{tabular}
\label{table:testing_rewards} 
\end{table}\\
Table ~\ref{table:testing_rewards} shows the average rewards obtained by the agents of both algorithms during testing. The process for calculating these reward values is different when compared to the training. During training, the average reward per episode was calculated. This table shows the average reward obtained during 1000 time steps the algorithm was run. It is interesting to note that both algorithms have similar values for each environment, while the D3QN-SAM had fewer crashes compared to the D3QN-CNN. One way to explain this is that the former algorithm learned a better and robust policy while obtaining the same rewards and further investigation needs to be done regarding this. The rewards obtained by the Complex Environment is slightly lower than that of the Basic Environment. This can be attributed to the fact that it has many obstacles present in the form of vertical beams and one aspect of the reward depends on how far the agent is from any obstruction. Furthermore, the rewards present in the Turbulent Environment seem inconsistent with the other environments. The objects floating around in the Turbulent Environment, such as leaves, cause the depth sensor to obtain low values of depth even when the agent is far from walls or beams. These low values in turn generate negative reward values.
\\

\subsection{Interpreting the D3QN-SAM}
\begin{figure}[!ht]
\begin{subfigure}{0.45\columnwidth}
  \centering
  \includegraphics[width=3.5 cm]{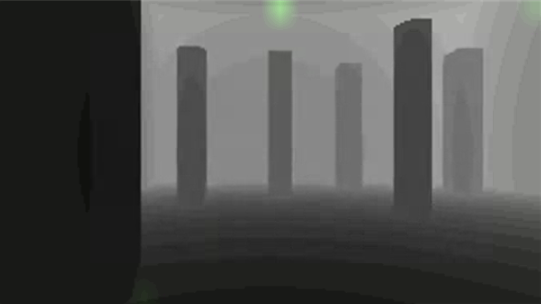}
  \caption{}
  \label{fig:sfig1}
\end{subfigure}%
\begin{subfigure}{0.45\columnwidth}
  \centering
  \includegraphics[width=3.5 cm]{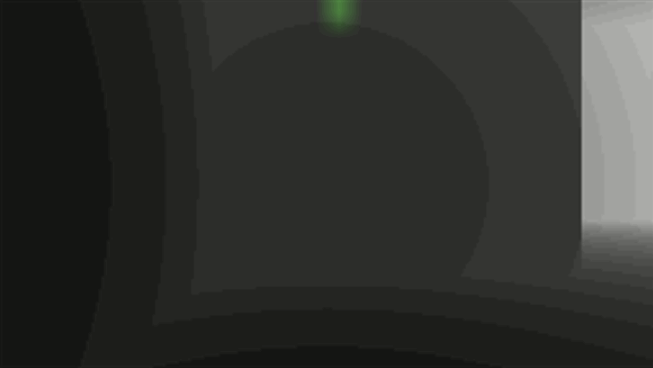}
  \caption{}
  \label{fig:sfig2}
\end{subfigure}
\caption{Visualizing Attention Weights in a) Turbulent and b) Basic Environment}
\label{fig:attention_weight}
\end{figure}

A major use of attention models in DNN is the ability to visualize its weights, enabling a better understanding of how the network makes its decisions. The attention weights of the D3QN agent has been visualized, presented in green color, while it was placed in the Basic and Turbulent Environment (Figure~\ref{fig:attention_weight}). It can be seen that a relatively large amount of attention is placed on walls that are farthest from the UAV, which can be interpreted as the agent trying to focus on the optimal distance to increase it rewards. However, the choice to place the attention weights in Figure~\ref{fig:sfig2} is sub-par, for it would have been better to focus on the opening next to the walls. Perhaps increasing the rewards for staying away from walls and obstacles would lead to better placement of attention weights \cite{zai2020deep}.
\section{Conclusion and Future Work}
This paper investigates the integration of SAM with DRL for the purpose of UAV Collision Avoidance, while comparing it with a conventional CNN based model. While the training results were quite similar, the DRL-SAM outperformed the DRL-CNN during testing. Also, we show that the DRL-SAM algorithm is interpretable and therefore can be incorporated into safety-critical vision algorithms.\\
The next steps will involve getting the algorithm ready to be transferred onto a real drone. To further reduce the chances of collisions both in simulation and real-life we propose, that along with an penalty of coming close to objects, the episode should end with a large negative reward if the depth values reduce below a threshold. This would further encourage the UAV to keep a minimum distance from obstacles under all circumstances. Finally, based on the work by \cite{singla2019memory} and \cite{hausknecht2015deep}, we suggest a study on understanding the efficacy of stacking a larger number of observations as input to a CNN. While traversing difficult environments, the UAV would need to memorize larger trajectories for cases where they are moving obstacles present. \\

\printbibliography

\end{document}